\pdfoutput=1

\documentclass[11pt]{article}

\usepackage[]{ACL2023}

\usepackage{times}
\usepackage{latexsym}

\usepackage[T1]{fontenc}

\usepackage[utf8]{inputenc}

\usepackage{microtype}

\usepackage{inconsolata}

\usepackage{bm}
\usepackage{amsmath}
\usepackage{url}
\usepackage{graphicx}
\usepackage{float} 
\usepackage{multirow}
\usepackage{makecell}
\usepackage{xpinyin}
\usepackage{enumerate}

\usepackage{xcolor}
\usepackage{framed}
\usepackage{color}
\usepackage{bbding}
\usepackage{soul}
\usepackage{subcaption}
\usepackage{colortbl}
\newcolumntype{L}[1]{>{\raggedright\let\newline\\\arraybackslash\hspace{0pt}}m{#1}}
\setul{2pt}{2pt}
\usepackage{booktabs}
\usepackage{amsmath,amsfonts}
\usepackage[utf8]{inputenc}
\usepackage{cleveref}

\usepackage{ulem}
\usepackage{diagbox}
\crefname{section}{§}{§§}
\Crefname{section}{§}{§§}
\usepackage{enumitem}
\setenumerate[1]{itemsep=0pt,partopsep=0pt,parsep=\parskip,topsep=5pt}
\setitemize[1]{itemsep=0pt,partopsep=0pt,parsep=\parskip,topsep=5pt}
\setdescription{itemsep=0pt,partopsep=0pt,parsep=\parskip,topsep=5pt}
\usepackage{arydshln}

\usepackage[linesnumbered,ruled,vlined]{algorithm2e}

\title{Unified Model Learning for Various Neural Machine Translation}

\author{
 Yunlong Liang$^{\spadesuit}$\footnotemark[2]\hspace{0.6mm}, 
 Fandong Meng$^{\heartsuit}$\footnotemark[3]\hspace{0.6mm}, 
 Jinan Xu$^{\spadesuit}$\footnotemark[3]\hspace{0.6mm}, 
 Jiaan Wang$^{\heartsuit}$\hspace{0.2mm},
 Yufeng Chen$^{\spadesuit}$\hspace{0.5mm}
 and {Jie Zhou}$^{\heartsuit}$\hspace{0.2mm}\hspace{1.5mm}\\
 $^\spadesuit$ Beijing Key Lab of Traffic Data Analysis and Mining, \\Beijing Jiaotong University, Beijing, China \\
 $^\heartsuit$ Pattern Recognition Center, WeChat AI, Tencent Inc, China\\
  \texttt{\{yunlongliang,jaxu\}@bjtu.edu.cn} \\
  \texttt{fandongmeng@tencent.com} \\
}

\begin{document}
\begin{CJK}{UTF8}{gkai}
\maketitle
\renewcommand{\thefootnote}{\fnsymbol{footnote}}
\footnotetext[2]{Work was done when Yunlong was interning at Pattern Recognition Center, WeChat AI, Tencent Inc, China.}
\footnotetext[3]{Corresponding authors.}
\renewcommand{\thefootnote}{\arabic{footnote}}

\begin{abstract}
Existing neural machine translation (NMT) studies mainly focus on developing dataset-specific models based on data from different tasks (\emph{e.g.}, document translation and chat translation). Although the dataset-specific models have achieved impressive performance, it is cumbersome as each dataset demands a model to be designed, trained, and stored. In this work, we aim to unify these translation tasks into a more general setting. Specifically, we propose a ``versatile'' model, \emph{i.e.}, the \textbf{U}nified \textbf{M}odel \textbf{L}earning for \textbf{NMT} ({\fontfamily{lmtt}\selectfont UMLNMT}) that works with data from different tasks, and can translate well in multiple settings simultaneously, and theoretically it can be as many as possible. Through unified learning, {\fontfamily{lmtt}\selectfont UMLNMT} is able to jointly train across multiple tasks, implementing intelligent on-demand translation. On seven widely-used translation tasks, including sentence translation, document translation, and chat translation, our {\fontfamily{lmtt}\selectfont UMLNMT} results in substantial improvements over dataset-specific models with significantly reduced model deployment costs. Furthermore, {\fontfamily{lmtt}\selectfont UMLNMT} can achieve competitive or better performance than state-of-the-art dataset-specific methods. Human evaluation and in-depth analysis also demonstrate the superiority of our approach on generating diverse and high-quality translations. Additionally, we provide a new genre translation dataset about famous aphorisms with 186k Chinese$\rightarrow$English sentence pairs.


\end{abstract}

\section{Introduction}
Neural machine translation (NMT) tasks, including sentence translation~\cite{zhangetal2019bridging}, document translation~\cite{maruf-etal-2019-selective}, chat translation~\cite{farajian-etal-2020-findings,liang-etal-2022-msctd}, personalized translation~\cite{lin-etal-2021-towards}, multimodal translation~\cite{elliott-etal-2016-multi30k}, and domain-specific translation~\cite{bawden-etal-2019-findings,bawden-etal-2020-findings}, have received considerable attention in recent years. According to their different task definitions, previous research on each task mainly focuses on designing dataset-specific architectures and objectives, having obtained superior performance. 
\textbf{\begin{figure*}[t]
    \centering
    \includegraphics[width=0.99\textwidth]{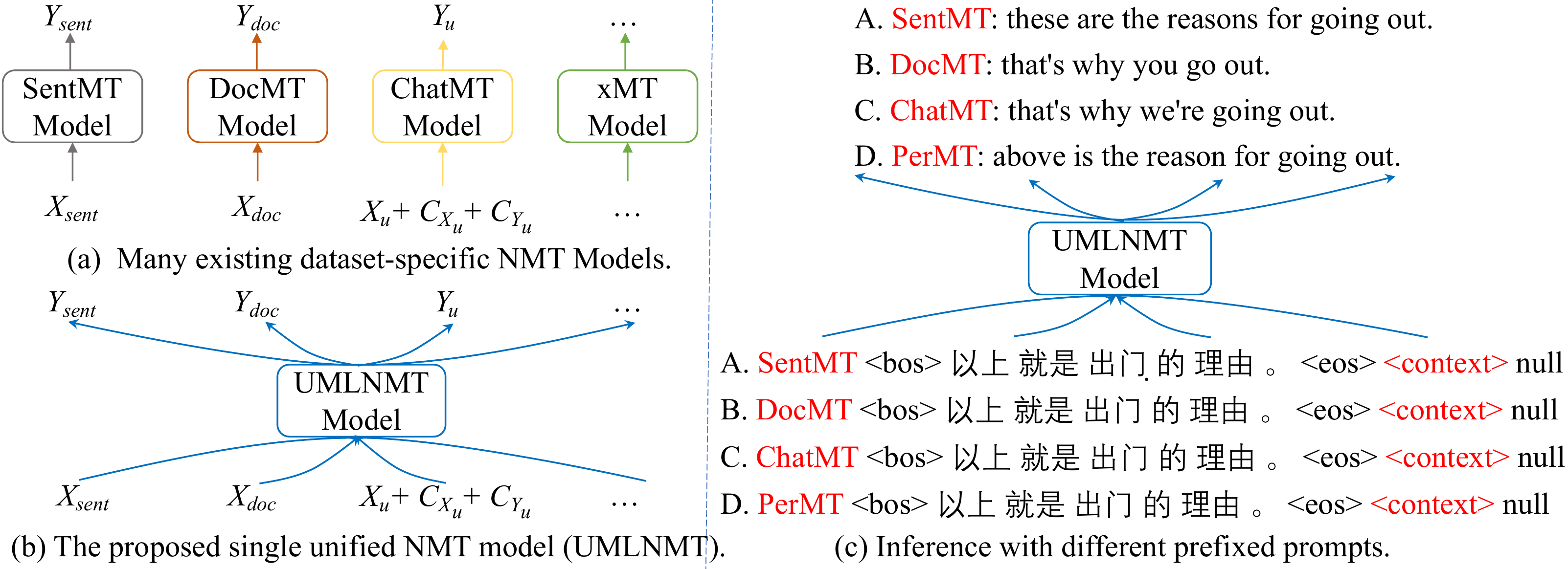}
    \caption{Comparison of (a) existing dataset-specific NMT models and (b) the proposed unified one. (c) During inference, our {\fontfamily{lmtt}\selectfont UMLNMT} is able to generate diverse and high-quality translations for the same input with different prefixed prompts. ``SentMT'', ``DocMT'', ``ChatMT'', and ``PerMT'' denote sentence, document, chat, and personalized machine translation, respectively. }
    \label{fig.1}\vspace{-5pt}
\end{figure*}}

Intuitively, there is a close relationship among these translation tasks because they all require models to translate the given text. Unfortunately, existing work only studies each task separately, as shown in~\autoref{fig.1} (a), which would require a large amount of computation resource to train so many models and deployment costs. Besides, each of them is still a single dataset-specific model capable of translating well on a setting, rather than a ``versatile'' model that can handle multiple settings simultaneously. This is frustrating in practice: for example, an NMT model trained on a medical dataset can translate entities well related to the medical domain such as ``COVID-19'' but it may not work for entities from other domains such as ``Stock-for-Stock'' in the financial field. As a result, such NMT methods can not scale up well as each dataset requires a model to be trained and stored.  

Apparently, building a versatile model to handle all scenes is more attractive and much-needed. Furthermore, a unified model would achieve better performance than dataset-specific models if we can make the most of these different datasets. Especially for some limited scenarios, the paired dataset is scarce and costly to collect (\emph{e.g.}, only 17k triplet data are available for personalized translation~\cite{lin-etal-2021-towards}). What's more, when applying an NMT model to one specific text, users are not always interested in a fixed output~\cite{susanto-etal-2020-lexically,ijcai2020p496,Chen_Chen_Li_2021,ijcai2021p547}. However, existing NMT systems only generate one fixed translation for the same input, which is not an ideal delivery mode. All of the above call for an on-demand and versatile NMT model that not only supports multiple translation setups but also flexibly produces requested translation types. 

In this work, we propose to unify these NMT tasks into a more general setting and present a versatile model, namely, {the \textbf{U}nified \textbf{M}odel \textbf{L}earning for  \textbf{NMT} ({\fontfamily{lmtt}\selectfont UMLNMT})}, which can handle multiple NMT settings simultaneously, as shown in~\autoref{fig.1} (b). {\fontfamily{lmtt}\selectfont UMLNMT} is built upon the publicly used transformer backbone~\cite{vaswani2017attention} and empowered by unified learning. Then, we jointly train the model on six and seven different datasets in Chinese$\rightarrow$English (Zh$\rightarrow$En) and English$\rightarrow$German (En$\rightarrow$De) directions, respectively. An obvious benefit of a unifying-based design is that the prompts can be served as instructions to guide the model to provide diverse outputs depending on the required types of interest to the user, examples of which are shown in the red prompt text on the right part of~\autoref{fig.1} (c). 

We validate our {\fontfamily{lmtt}\selectfont UMLNMT} model on 7 types of translation tasks, including WMT20 news translation, document translation, chat translation, personalized translation, multimodal translation, domain-specific translation (\emph{e.g.}, biomedical translation), and aphorism translation (on a self-collected dataset), involving Zh$\rightarrow$En and En$\rightarrow$De directions. Extensive experiments show that {\fontfamily{lmtt}\selectfont UMLNMT} achieves significantly better performance than models of the same architecture trained with a single dataset. This is a promising result because different NMT datasets vary greatly in the task definition, data format, corpus domain, and scale, and it is challenging to have a single versatile model that is able to handle multiple scenarios simultaneously~\cite{lu202012,li2022grounded,kamath2021mdetr}. Additionally, {\fontfamily{lmtt}\selectfont UMLNMT} gains competitive or better results than the state-of-the-art dataset-specific models in terms of BLEU~\cite{papineni2002bleu}, TER~\cite{snover2006study}, ChrF2~\cite{popovic-2015-chrf}, COMET~\cite{rei-etal-2020-comet}, and BLEURT~\cite{sellam-etal-2020-bleurt} scores, showing its superiority and generalizability. Human evaluation and in-depth analysis also suggest that our {\fontfamily{lmtt}\selectfont UMLNMT} can produce diverse and fluent translations. 

Our contributions are summarized as follows~\footnote{The data will be released at: \url{https://github.com/XL2248/UMLNMT}. }:

\begin{itemize}[leftmargin=*]

\item To the best of our knowledge, we are the first that proposes a unified NMT benchmark setting, \emph{i.e.}, unifying multiple NMT tasks to a more general setting. We present {\fontfamily{lmtt}\selectfont UMLNMT}, a new method that supports translation in many scenarios in a single model. We also show that given different prompts, the model can generate diverse translations of interest to the user, enabling on-demand translation. 

 \item We demonstrate that a single model works well on the unified benchmark. This suggests our {\fontfamily{lmtt}\selectfont UMLNMT} could replace a series of dataset-specific models, saving a lot of parameters. Experiments show that our {\fontfamily{lmtt}\selectfont UMLNMT} is able to benefit from different tasks and datasets and thus achieves substantially better performance than dataset-specific models. 

\item We contribute a new genre translation dataset about famous aphorisms with 186k Zh$\rightarrow$En sentence pairs to the research community. Besides, we will release a test set (1400 instances) that is annotated with detailed mis-translation, over-translation, under-translation, and grammatical errors.
  
\end{itemize}

\section{Background}
\textbf{Sentence Machine Translation (SentMT).}
Given an input sentence in the source language $X_{sent}$$=$$\{x_i\}_{i=1}^{|X_{sent}|}$, the goal of the SentMT model is to produce its translation in the target language $Y_{sent}$$=$$\{y_i\}_{i=1}^{|Y_{sent}|}$. The conditional distribution of the model is:
\begin{equation}\nonumber
\setlength{\abovedisplayskip}{8pt}
\setlength{\belowdisplayskip}{8pt}
\label{eq:nmt}
    p_{\theta}(Y_{sent}|X_{sent}) = \prod_{t=1}^{|Y_{sent}|}p_{\theta}(y_t|X_{sent}, y_{1:t-1}),
\end{equation}
where $\theta$ are model parameters and $y_{1:t-1}$ is the partial translation. 

\noindent\textbf{Document Machine Translation (DocMT).}
Given an input document in the source language $X_{doc}$$=$$\{X_{sent}^{i}\}_{i=1}^{M}$ and the corresponding target document in the target language $Y_{doc}$$=$$\{Y_{sent}^{i}\}_{i=1}^{M}$, following~\citet{zhang-etal-2018-improving}, the document machine translation can be approximated as:
\begin{equation}\nonumber
\setlength{\abovedisplayskip}{8pt}
\setlength{\belowdisplayskip}{8pt}
\label{eq:ms}
    p_{\theta}(Y_{doc}|X_{doc}) \approx \prod_{i=1}^{M}p_{\theta}(Y_{sent}^{i}|X_{sent}^{i}, X_{doc}^{<i}),
\end{equation}
where $X_{doc}^{<i}$ is the document-level context used to
help translate $X_{sent}^{i}$. 

\noindent\textbf{Chat Machine Translation (ChatMT). }
The ChatMT task aims to generate $Y_u$=$\{y_{u,1}, y_{u,2}, ..., y_{u,T}\}$ with the guidance of the $u$-th utterance $X_u$=$\{x_{u,1}, x_{u,2}, ..., x_{u,N}\}$ and the associated bilingual dialogue history $\mathcal{C}_{X_u}$ and $\mathcal{C}_{Y_u}$. Formally, the probability distribution of the target utterance $Y_u$ is defined as follows:
\begin{equation}\nonumber
\label{eq:dnmt}
\setlength{\abovedisplayskip}{8pt}
\setlength{\belowdisplayskip}{8pt}
\resizebox{1.01\hsize}{!}{$
\begin{split}
        {p_{\theta}}(Y_{u}|X_u, \mathcal{C}_{X_u}, \mathcal{C}_{Y_u}) = \prod_{t=1}^{T}p_{\theta}(y_{u,t}|y_{u,<t}, X_{u}, \mathcal{C}_{X_u}, \mathcal{C}_{Y_u}),
\end{split}
$}
\end{equation}
where $y_{u,<t}$ = $\{y_{u,1}, y_{u,2}, y_{u,3}, ..., y_{u,t-1}\}$.

\noindent\textbf{Personalized Machine Translation (PerMT). }
Given the paired inputs <$X_{sent}^{p}$, $H^{p}$> where $H^{p}$ is the historical inputs of the user $p$ (\emph{e.g.}, topic preference, stylistic characteristics, and expression habit), the goal of PerMT is to learn a model that can generate a translation $Y_{sent}^{p}$ which can reflect the traits of user $p$. Formally, it is as follows: 
\begin{equation}\nonumber
\setlength{\abovedisplayskip}{8pt}
\setlength{\belowdisplayskip}{8pt}
\label{eq:permt}
    p_{\theta}(Y_{sent}^{p}|X_{sent}^{p}) = \prod_{t=1}^{|Y_{sent}^{p}|}p_{\theta}(y_t|X_{sent}^{p}, H^{p}, y_{1:t-1}).
\end{equation}

\noindent\textbf{Multimodal Machine Translation (MMT). }
Given the paired inputs <$X_{sent}$, $X_{img}$> where $X_{img}$ is usually the extracted image feature, the MMT aims to learn a model that can generate a translation $Y_{sent}$. Formally, it is as follows: 
\begin{equation}\nonumber
\setlength{\abovedisplayskip}{8pt}
\setlength{\belowdisplayskip}{8pt}
\resizebox{1.01\hsize}{!}{$
\begin{split}
\label{eq:mmt}
    p_{\theta}(Y_{sent}|X_{sent}) = \prod_{t=1}^{|Y_{sent}|}p_{\theta}(y_t|X_{sent}, X_{img}, y_{1:t-1}).
\end{split}
$}
\end{equation}

\noindent\textbf{Domain-specific Machine Translation (DsMT).}
Its formulation is the same as SentMT while the difference is the training data belonging to obviously distant domains (\emph{e.g.}, TED and News).

\noindent\textbf{Aphorism Machine Translation (AphMT).}
Its formulation is the same as SentMT while the difference is the training data are bilingual aphorisms.

\section{{\fontfamily{lmtt}\selectfont UMLNMT}}

Our {\fontfamily{lmtt}\selectfont UMLNMT} is based on the popular transformer backbone~\cite{vaswani2017attention}. We first present the model architecture and describe how we reframe the different NMT tasks into a unified general setting with prompts. We then present the datasets used in our experiments. Finally, we detail the training and inference process.

\subsection{Model Architecture}
\label{sec:enc}
Our model is based on the vanilla transformer encoder-decoder architecture \cite{vaswani2017attention}. We list main configurations in~\autoref{tab:train-detail} of the appendix. 

\begin{table*}[]
\small
\centering
\begin{tabular}{l|p{13.8cm}l}
\hline
 Source &以上 就是 出门 的 理由 。(\textcolor{red}{The pinyin style of Chinese:} yǐshàng jiùshì chūmén de lǐyóu) \\
 \hline
  Reference &All these were the reasons to go out. \\
 \hline
w/ SentMT &<SentMT> <bos> 以上 就是 出门 的 理由 。<eos> <context> NULL \\ \hline
\multirow{9}{*}{\rm{w/ DocMT}} &<DocMT> <bos> 以上 就是 出门 的 理由 。 <eos> <context> 打 起 点 精神 ， 对 狗狗纳纳 说 ： 妈妈 带 你 出门 。<sep>
细细 地 写 了 出门 要 办 的 几 宗 事情 ... <sep> 取 西服 ， 买 营养 粒 ， 付款 ， 买 生 鱼片 ， 纳纳 的 小 零食 。\\
&\textcolor{red}{The pinyin style of Chinese:}<DocMT> <bos> yǐshàng jiùshì chūmén de lǐyóu <eos> <context> dǎ qǐ diǎn jīngshén, duì gǒugǒunànà shuō: māmā dài nǐ chūmén. <sep> xìxì dì xiě le chūmén yào bàn de jǐ zōng shìqing... <sep> qǔ xīfú, mǎi yíngyǎng lì, fùkuǎn, mǎi shēng yúpiàn, nànà de xiǎo língshí.\\
&\textcolor{red}{English:} <DocMT> <bos> All these were the reasons to go out. <eos> <context> Pulling myself together, I said to my dog Nana, Mom's taking you out. <sep> I wrote down a few things I had to do when I left home... <sep> Get a suit, buy nutritional grain, pay my bills, and buy sashimi and snacks for Nana.\\
\hline
\end{tabular}
\caption{The example with different prefixed prompts.}
\label{example}\vspace{-13pt}
\end{table*}

\subsection{Task Reframing}

We reframe multiple NMT as a general unifying-based sequence-to-sequence task. Formally, given an original input sequence $X$ and the additional input $X_{add}$~\footnote{Note that $X_{add}$ is null if no context, \emph{e.g.}, SentMT.}, we transform $X$ and $X_{add}$ to a new sequence $X_{input}$ by prefixing it with a set of prompts as follows:
\begin{equation} \label{eq:prefix_x}
\setlength{\abovedisplayskip}{5pt}
\setlength{\belowdisplayskip}{5pt}
\begin{split}
X_{input} & = [s_{p},s_{bos},X, s_{eos}, s_{ctx}, X_{add}],
\end{split}
\end{equation}
\noindent where $X_{input}$ is the model input; $s_{p}$ is the specific translation type that we are interested in, \emph{e.g.}, ``sentMT''; $s_{bos}$ and $s_{eos}$ are the special tokens that indicate the beginning and the ending of the to be translated source sequence, respectively; $s_{ctx}$ is the special token \textit{``<context>''} indicating that the following sequence is the additional context of the current source sequence (it can be chat history or user traits or image features, etc). Note that $X_{add}$ may contain a special token $s_{sep}$ that delimits its included document context or dialogue history or historical inputs. Then the target sequence $Y_{output}$ is:
\begin{equation} \label{eq:prefix_y}\nonumber
\setlength{\abovedisplayskip}{5pt}
\setlength{\belowdisplayskip}{5pt}
\begin{split}
Y_{output} & = [s_{p}:Y],
\end{split}
\end{equation}
\noindent where $s_{p}$ is the desired translation type that are identical to those in~\autoref{eq:prefix_x}, and $Y \in \{Y_{sent}, Y_{doc}, Y_{u}, Y_{sent}^p\}$.   

For example, as shown in~\autoref{example}, suppose we have an input sentence $X$ \textit{``yǐshàng jiùshì chūmén de lǐyóu''} in SentMT. Then $X_{input}$ will be \textit{``<SentMT> <bos>yǐshàng jiùshì chūmén de lǐyóu<eos> <context> NULL''} and $Y_{output}$ will be \textit{``<SentMT>: All these were the reasons to go out.''}. Alternatively, if we have a context before the input in DocMT, then the $X_{input}$ will be \textit{``<DocMT> <bos> yǐshàng jiùshì chūmén de lǐyóu <eos> <context> dǎ qǐ diǎn jīngshén, duì gǒugǒunànà shuō: māmā dài nǐ chūmén. <sep> xìxì dì xiě le chūmén yào bàn de jǐ zōng shìqing... <sep> qǔ xīfú, mǎi yíngyǎng lì, fùkuǎn, mǎi shēng yúpiàn, nànà de xiǎo língshí.''} and $Y_{output}$ will be \textit{``<DocMT>: All these were the reasons to go out.''}. It is similar to other prompts. In this way, we hope prompts can serve as indicators that steer the model to generate the expected translation type.  

\subsection{Diverse NMT Datasets} \label{subsec:datasets}

We explore seven types of translation tasks involving the following datasets and we also list detailed statistics of them in~\autoref{tbl:data-statistics} of the Appendix.

\paragraph{SentMT.}
We use WMT2020 news translation dataset in Zh$\rightarrow$En and En$\rightarrow$De directions\footnote{https://www.statmt.org/wmt20/translation-task.html. }. Generally, we filter out duplicate sentence pairs and remove those whose length (character for Chinese, and word for English and German) exceeds 80. Then, we conduct full-/half-width conversion, unicode conversion, punctuation normalization, and tokenization. We take the newstest2019 as the development set and the newstest2020 as the test set.

\paragraph{DocMT.}
Following~\citet{maruf-etal-2019-selective,DBLP:journals/corr/abs-2002-07982,sun-etal-2022-rethinking}, we use these datasets:
\begin{itemize}[leftmargin=*]
\item \noindent {TED (Zh$\rightarrow$En/En$\rightarrow$De)}. The Zh$\rightarrow$En and En$\rightarrow$De TED datasets are from IWSLT 2015 and 2017 evaluation campaigns, respectively. For Zh$\rightarrow$En, we take dev2010 as the development set and the merged tst2010-2013 as the test set. For En$\rightarrow$De, we use the merged tst2016-2017 as our test set and the rest as the development set.

\item \noindent {Subtitle (Zh$\rightarrow$En)}. We use the subtitle data in~\citet{liang-etal-2022-scheduled} and we randomly split the training, development, and test set.

\item \noindent {News (En$\rightarrow$De)}. We use News Commentary v11 as our training set. The WMT newstest2015 and newstest2016 are used as the development set and test set, respectively.

\item \noindent {Europarl (En$\rightarrow$De)}. We follow the method~\cite{maruf-etal-2019-selective} to extract the training, development, and test set from the Europarl v7.

\end{itemize} 

\paragraph{ChatMT.} 
We utilize the human-annotated bilingual dialogue MSCTD dataset~\cite{liang-etal-2022-msctd} for chat machine translation. For Zh$\rightarrow$En, it contains 10,749 training/504 development/509 test dialogues. For En$\rightarrow$De, it contains 2,066 training/504 development/509 test dialogues.

\paragraph{PerMT.}
For personalized machine translation, we leverage the human-annotated Zh$\rightarrow$En UDT-Corpus~\cite{lin-etal-2021-towards} containing 57,639 inputs of 6,550 users. Specifically, it includes 33,441, 3,629, and 3,470 historical inputs for training, development, and test sets, respectively.  

\paragraph{MMT.}
For multimodal machine translation, following previous work~\cite{yin-etal-2020-novel,fang-feng-2022-neural}, we utilize the Multi30K~\cite{elliott-etal-2016-multi30k} dataset containing En$\rightarrow$De sentence pairs with image annotations. We also report the results on the WMT17 test set and the ambiguous MSCOCO test set, which contain 1,000 and 461 instances, respectively. The image feature is obtained using an off-shelf Faster R-CNNs~\cite{NIPS2015_14bfa6bb} pre-trained on Visual Genome~\cite{krishnavisualgenome}. Specifically, for an image, we obtain a set of detected objects from Faster R-CNNs, \emph{i.e.}, $\mathbf{X}_{img}$ = $\{\mathbf{o}_{j,1}, \mathbf{o}_{j,2}, \mathbf{o}_{j,3}, ..., \mathbf{o}_{j,m}\}$, where $m$ is the number of extracted objects and $\mathbf{o}_{j,*} \in \mathbb{R}^{d_f}$.

\paragraph{DsMT.}
For domain-specific machine translation, we follow~\cite{https://doi.org/10.48550/arxiv.2209.11409} and use 4 datasets (Law, Medical, Koran and IT) proposed by~\citet{koehn-knowles-2017-six} and re-splited by~\citet{aharoni-goldberg-2020-unsupervised}. 

\paragraph{AphMT.}
We collect a bilingual aphorism dataset from the website\footnote{https://www.jiemengz.com/} for aphorism translation in Zh$\rightarrow$En direction. Specifically, we split it into 181,451 training/2,500 development/2,500 test sets. 

\subsection{Training and Inference}
At training, we randomly select instances from different datasets for each batch. And we train the model with the cross-entropy loss.

During inference, we generate translations by using all prefixed prompts for each input. For example, for an input sequence, we have five $s_p$ in~\autoref{eq:prefix_x} for both Zh$\rightarrow$En and En$\rightarrow$De as the prefixed prompts. Therefore, we can obtain five translations for each input via one single unified model, as shown in~\autoref{fig.1} (c).

\section{Experiments}
\subsection{Metrics}
For a fair comparison, we follow previous work~\cite{lin-etal-2021-towards,sun-etal-2022-rethinking,liang-etal-2021-towards} and adopt lexical-based metrics, \emph{e.g.}, SacreBLEU\footnote{BLEU+case.mixed+numrefs.1+smooth.exp+tok.13a+\\version.1.4.13}~\cite{post-2018-call}, ChrF2~\cite{popovic-2015-chrf} and TER~\cite{snover2006study} with the statistical significance test~\cite{koehn-2004-statistical}. Specifically, we report the case-insensitive BLEU score for Zh$\rightarrow$En, and the case-sensitive BLEU score for En$\rightarrow$De. Besides, we utilize some recent state-of-the-art evaluation metrics that highly correlate with human judgment, \emph{e.g.}, BLEURT~\cite{sellam-etal-2020-bleurt} and COMET~\cite{rei-etal-2020-comet}.

\subsection{Implementation Details}
In this paper, we train all models using standard transformer~\cite{vaswani2017attention} with \emph{base} and \emph{big} settings. We list our training details in~\autoref{ImD}.

\begin{table*}[t]
\begin{subtable}{.98\textwidth}
\centering
\scalebox{0.86}{
\setlength{\tabcolsep}{0.5mm}{
\begin{tabular}{l|l|cc|cc|c|c|c|cc}
\hline
&\multirow{2}{*}{Methods} &\multicolumn{2}{c|}{\textbf{SentMT}} &\multicolumn{2}{c|}{\textbf{DocMT}} &\multicolumn{1}{c|}{\textbf{ChatMT}} &\multicolumn{1}{c|}{\textbf{PerMT}} &\multicolumn{1}{c|}{\textbf{AphMT}} &\multicolumn{2}{c}{Overall} \\ 
\cline{3-6} \cline{7-11}
&\multicolumn{1}{l|}{} & \multicolumn{1}{c}{test19} & \multicolumn{1}{c|}{test20} & \multicolumn{1}{c}{TED} & \multicolumn{1}{c|}{subtitle} &  \multicolumn{1}{c|}{chat}      & \multicolumn{1}{c|}{personalized} & \multicolumn{1}{c|}{aphorism} & \multicolumn{1}{c}{Avg.} & \multicolumn{1}{c}{\# Params}  \\ \hline
\multirow{2}{*}{\rm{BLEU}$\uparrow$}
&SDTM  &25.56 &27.49  &24.96 &\textbf{30.90} &32.45 &36.13  &42.88 &31.48 &$N$*103M \\
&{\fontfamily{lmtt}\selectfont UMLNMT} &\;\textbf{30.64}$^{\dagger}$ &\;\textbf{30.75}$^{\dagger}$	&\;\textbf{25.60}$^{\dagger}$	&{30.52} &\;\textbf{36.93}$^{\dagger}$  &\;\textbf{37.73}$^{\dagger}$      &\;\textbf{43.46}$^{\dagger}$ &\;\textbf{33.66}$^{\dagger}$&1*103M \\\hline
\multirow{2}{*}{\rm{BLEURT}$\uparrow$}
&SDTM  &56.84 &63.40  &66.27 &65.13&67.00 &63.91  &73.72 &65.18 &$N$*103M\\
&{\fontfamily{lmtt}\selectfont UMLNMT} &\;\textbf{65.36}$^{\dagger}$ &\;\textbf{64.65}$^{\dagger}$	&\;\textbf{66.99}$^{\dagger}$	&\textbf{65.35}&\;\textbf{68.62}$^{\dagger}$  &\textbf{64.19}      &\;\textbf{74.60}$^{\dagger}$  &\;\textbf{67.11}$^{\dagger}$ &1*103M\\
\hline
\end{tabular}}}
\caption{Zh$\rightarrow$En. SDTM: Single-Dataset Training Model.}
\end{subtable}
\begin{subtable}{.980\textwidth}
\centering
\scalebox{0.675}{
\setlength{\tabcolsep}{0.5mm}{
\begin{tabular}{l|l|cc|ccc|ccc|c|cccc|cc}
\hline
&\multirow{2}{*}{Methods} &\multicolumn{2}{c|}{\textbf{SentMT}} &\multicolumn{3}{c|}{\textbf{DocMT}} &\multicolumn{3}{c|}{\textbf{MMT}} &\multicolumn{1}{c|}{\textbf{ChatMT}} &\multicolumn{4}{c|}{\textbf{DsMT}} &\multicolumn{2}{c}{{Overall}}\\ 
\cline{3-6} \cline{7-17}
&\multicolumn{1}{l|}{} & \multicolumn{1}{l}{test19} & \multicolumn{1}{l|}{test20} & \multicolumn{1}{c}{TED} & \multicolumn{1}{c}{News} & \multicolumn{1}{c|}{Europarl} &  \multicolumn{1}{c}{test2016}      & \multicolumn{1}{c}{test2017} & \multicolumn{1}{c|}{MSCOCO} & \multicolumn{1}{c|}{chat}& \multicolumn{1}{c}{Law} & \multicolumn{1}{c}{Medical} & \multicolumn{1}{c}{Koran} & \multicolumn{1}{c|}{IT} & \multicolumn{1}{c}{Avg.} & \multicolumn{1}{c}{\# Params}\\ \hline
\multirow{2}{*}{\rm{BLEU}$\uparrow$}
&SDTM  &40.53 &31.05 &27.86 &30.90 &31.11 &40.82&36.67 &\textbf{33.72}&\textbf{54.59}&30.72&28.65&30.72&28.65&35.14&$N$*85M\\
&{\fontfamily{lmtt}\selectfont UMLNMT} &\;\textbf{41.42}$^{\dagger}$	&\;\textbf{32.83}$^{\dagger}$	&\;\textbf{28.82}$^{\dagger}$	&\;\textbf{35.76}$^{\dagger}$  &\textbf{31.24}     &\textbf{41.44} &\;\textbf{38.10}$^{\dagger}$  &31.91&52.77&\;\textbf{32.03}$^{\dagger}$&\;\textbf{30.85}$^{\dagger}$&\;\textbf{32.03}$^{\dagger}$&\;\textbf{30.85}$^{\dagger}$&\;\textbf{36.11}$^{\dagger}$&1*85M\\\hline
\multirow{2}{*}{\rm{BLEURT}$\uparrow$}
&SDTM &68.39 & 64.97 &71.28  & 72.66 &\textbf{77.36}  & 74.08& 72.33 & \textbf{69.88}& \textbf{78.31}& 65.04& 63.36& 65.04& 63.36&70.69&$N$*85M\\
&{\fontfamily{lmtt}\selectfont UMLNMT} & \;\textbf{70.12}$^{\dagger}$	& \;\textbf{66.79}$^{\dagger}$	&\;\textbf{72.23}$^{\dagger}$	& \textbf{72.92}  & 77.32      & \;\textbf{75.17}$^{\dagger}$ & \;\textbf{74.87}$^{\dagger}$  & 69.22& 77.53& \;\textbf{76.69}$^{\dagger}$& \;\textbf{74.90}$^{\dagger}$& \;\textbf{76.69}$^{\dagger}$& \;\textbf{74.90}$^{\dagger}$&\;\textbf{73.43}$^{\dagger}$&1*85M\\
\hline
\end{tabular}}}
\caption{En$\rightarrow$De. SDTM: Single-Dataset Training Model.}
\end{subtable}
\vspace{-3mm}
\caption{Comparison to single-dataset training models (SDTM) with BLEU and BLEURT scores (\%) in Zh$\rightarrow$En and En$\rightarrow$De directions on different test sets. The ``N'' indicates the number of dataset-specific models. ``$^{\dagger}$'' indicates that statistically significantly better than the ``SDTM'' with t-test {\em p} \textless \ 0.01.}
\label{mainres1}
\vspace{-5pt}
\end{table*}

\subsection{Comparison Models}
We compare with the following state-of-the-art dataset-specific methods: 

\noindent\textbf{Doc2Sent++}~\cite{sun-etal-2022-rethinking}. For DocMT, it proposes an effective training technique to train the vanilla Transformer where the additional sentence corpus is used (2 million WMT2019 and 2.4 million Wikipedia sentence pairs for Zh$\rightarrow$En and En$\rightarrow$De, respectively).

\noindent\textbf{CA-MCT}~\cite{liang-etal-2022-msctd}. For ChatMT, it first trains the model on the WMT2020 sentence corpus and then fine-tuned it on the MSCTD data. 

\noindent\textbf{UD-NMT}~\cite{lin-etal-2021-towards}. For PerMT, it first trains the model on the WMT2017 sentence corpus and then fine-tuned it on the UDT dataset.

\noindent\textbf{PLUVR}~\cite{fang-feng-2022-neural}.
This method uses the phrase-level universal visual representation for MMT to enhance the model.

\noindent\textbf{RePP}~\cite{fang-feng-2022-neural}.
For DsMT, this method directly trains their domain-specific model on the corresponding dataset.
\begin{table*}[t]
\begin{subtable}{.98\textwidth}
\centering
\scalebox{0.9}{
\setlength{\tabcolsep}{0.5mm}{
\begin{tabular}{l|cc|cc|c|c|c|c}
\hline
\multirow{2}{*}{Methods} &\multicolumn{2}{c|}{\textbf{SentMT}} &\multicolumn{2}{c|}{\textbf{DocMT}} &\multicolumn{1}{c|}{\textbf{ChatMT}} &\multicolumn{1}{c|}{\textbf{PerMT}} &\multicolumn{1}{c|}{\textbf{AphMT}} &\multirow{2}{*}{\# Models} \\ 
\cline{2-6} \cline{7-8}
\multicolumn{1}{l|}{} & \multicolumn{1}{c}{test19} & \multicolumn{1}{c|}{test20} & \multicolumn{1}{c}{TED} & \multicolumn{1}{c|}{subtitle}&  \multicolumn{1}{c|}{chat}      & \multicolumn{1}{c|}{personalized}  & \multicolumn{1}{c|}{aphorism} &\multicolumn{1}{l}{}  \\ \hline
Doc2Sent++~\cite{sun-etal-2022-rethinking}  & - &- &22.00  &-     &-  &- &- & 6  \\
CA-MCT~\cite{liang-etal-2022-msctd} & - &- &-  &-  &28.81      &- &- & 6  \\
UD-NMT~\cite{lin-etal-2021-towards}  & - &- & - &-  &-      &32.35  &- & 6  \\\cdashline{1-9}[4pt/2pt]
{\fontfamily{lmtt}\selectfont UMLNMT} (base) &30.64 &30.75	&\;25.60$^{\dagger}$	&30.52 &\;\textbf{36.93}$^{\dagger}$  &\;37.73$^{\dagger}$     &43.46 &1 \\
{\fontfamily{lmtt}\selectfont UMLNMT} (big)  &\textbf{31.10}  &\textbf{31.05} &\;\textbf{25.92}$^{\dagger}$  &\textbf{30.71}&\;{36.74}$^{\dagger}$ &\;\textbf{39.61}$^{\dagger}$    &\textbf{43.76}&1\\ 
\hline
\end{tabular}}}
\caption{Zh$\rightarrow$En. The ``-'' indicates no such result in the original paper. }
\end{subtable}
\begin{subtable}{.98\textwidth}
\centering
\scalebox{0.70}{
\setlength{\tabcolsep}{0.4mm}{
\begin{tabular}{l|cc|ccc|ccc|c|cccc|cc}
\hline
\multirow{2}{*}{Methods} &\multicolumn{2}{c|}{\textbf{SentMT}} &\multicolumn{3}{c|}{\textbf{DocMT}} &\multicolumn{3}{c|}{\textbf{MMT}} &\multicolumn{1}{c|}{\textbf{ChatMT}} &\multicolumn{4}{c|}{\textbf{DsMT}} &\multirow{2}{*}{\# Models}\\ 
\cline{2-6} \cline{7-14}
\multicolumn{1}{l|}{} & \multicolumn{1}{l}{test19} & \multicolumn{1}{l|}{test20} & \multicolumn{1}{c}{TED} & \multicolumn{1}{c}{News} & \multicolumn{1}{c|}{Europarl} &  \multicolumn{1}{c}{test2016}      & \multicolumn{1}{c}{test2017} & \multicolumn{1}{c|}{MSCOCO} & \multicolumn{1}{c|}{chat}& \multicolumn{1}{c}{Law} & \multicolumn{1}{c}{Medical} & \multicolumn{1}{c}{Koran} & \multicolumn{1}{c|}{IT} \\ \hline
Doc2Sent++~\cite{sun-etal-2022-rethinking}   & - &- &27.34&29.50&\textbf{32.44} & - &- & - &- & - & -& - & - &7\\
PLUVR~\cite{fang-feng-2022-neural}  & - &- & - &- & - &40.30 &33.45&30.28&-& - & - & - & -&7\\
CA-MCT~\cite{liang-etal-2022-msctd}   & - &- & - &- & - &- & - &- &52.72& - & - & - & - &7\\
RePP~\cite{https://doi.org/10.48550/arxiv.2209.11409}   & - &- & - &- & - &- & - &- &-&50.95 &47.48 &18.13 &39.57 &7\\\cdashline{1-15}[4pt/2pt]
{\fontfamily{lmtt}\selectfont UMLNMT} (base) &41.42	&32.83	&\;28.82$^{\dagger}$	&\;35.76$^{\dagger}$  &31.24     &\;41.44$^{\dagger}$ &\;38.10$^{\dagger}$ &\;\textbf{31.91}$^{\dagger}$&52.77&\textbf{32.03}&\textbf{30.85}&\textbf{32.03}&\textbf{30.85}&1\\
{\fontfamily{lmtt}\selectfont UMLNMT} (big)  &\textbf{42.03} &\textbf{33.11} &\;\textbf{29.33}$^{\dagger}$  &\;\textbf{36.17}$^{\dagger}$ &{31.57}  &\;\textbf{43.96}$^{\dagger}$  &\;\textbf{39.07}$^{\dagger}$&\;{31.17}$^{\dagger}$&\;\textbf{53.75}$^{\dagger}$&30.56&29.21&30.56&29.21&1\\
\hline
\end{tabular}}}
\caption{En$\rightarrow$De.  The ``-'' indicates no such result in the original paper. }
\end{subtable}
\vspace{-3mm}
\caption{Comparison to state-of-the-art models with BLEU scores (\%) in Zh$\rightarrow$En and En$\rightarrow$De directions on different test sets. ``$^{\dagger}$'' indicates that statistically significant better than compared models with t-test {\em p} \textless \ 0.01. }
\label{mainres2}
\vspace{-12pt}
\end{table*}

\subsection{Main Results}
\subsubsection{Comparison to Single-Dataset Performance} 
In this section, we aim to answer the question: \textbf{Can our multi-dataset jointly trained {\fontfamily{lmtt}\selectfont UMLNMT} can surpass the single-dataset trained models}?

Therefore, we conduct experiments in Zh$\rightarrow$En and En$\rightarrow$De directions, shown in~\autoref{mainres1}. Firstly, following~\citet{sun-etal-2022-rethinking,liang-etal-2022-msctd,lin-etal-2021-towards}, we employ the pretraining-then-fine-tuning paradigm, \emph{i.e.}, first pretraining the model on the WMT2020 sentence corpus and then continue training the model on each NMT dataset independently (for {\fontfamily{lmtt}\selectfont UMLNMT}, continue training on merged datasets). For single-dataset training models (SDTM), we select the best-performing model on a validation set and report its performance on the corresponding test set. For {\fontfamily{lmtt}\selectfont UMLNMT}, we select the model with the best averaged BLEU score on all NMT validation sets and report scores on each test set separately. 

\autoref{mainres1} presents the experimental results. It shows that the jointly trained {\fontfamily{lmtt}\selectfont UMLNMT} model outperforms SDTM by average improvement of 2.18 BLEU and 1.93 BLEURT points in Zh$\rightarrow$En direction, and 0.97 BLEU and 2.74 BLEURT points in En$\rightarrow$De direction. It suggests that our joint training on multiple datasets can mutually benefit each other rather than degrades each other. And our single model {\fontfamily{lmtt}\selectfont UMLNMT} significantly reduces $N$ times of model parameters than dataset-specific models. To be specific, in terms of BLEU scores on seven out of thirteen datasets, {\fontfamily{lmtt}\selectfont UMLNMT} outperforms the single-dataset trained counterparts by a large margin especially in test19 (Zh$\rightarrow$En, 30.64 vs. 25.56), chat (Zh$\rightarrow$En, 36.93 vs. 32.45) and News (35.76 vs. 30.90), and with some improvements in TED (Zh$\rightarrow$En, 25.60 vs. 24.96), aphorism (43.46 vs 42.88) and test2016 (41.44 vs. 40.82). In subtitle and Europarl, the performance of {\fontfamily{lmtt}\selectfont UMLNMT} is also comparable. The {\fontfamily{lmtt}\selectfont UMLNMT} fails to exceed SDTM on all three test sets of MMT (except MSCOCO) and another exception is chat translation (En$\rightarrow$De). This is probably because we select the checkpoint that is generally useful for all datasets according to the averaged performance on all validation sets other than on the validation set of MMT or chat. In the follow-up observations, we select the ``best'' checkpoint by the validation set performance of MMT and chat, respectively. The results on the test sets (MSCOCO and chat) are 33.79 and 54.65 BLEU scores respectively, which are comparable with SDTM. 

We list the results of ChrF2, TER, and COMET in~\autoref{tbl:main-res-others} of Appendix where the {\fontfamily{lmtt}\selectfont UMLNMT} still achieves better performance in most cases, showing its superiority. 

\begin{table*}[t!]
\centering
\newcommand{\tabincell}[2]{\begin{tabular}{@{}#1@{}}#2\end{tabular}}
\scalebox{0.85}{
\setlength{\tabcolsep}{0.5mm}{
\begin{tabular}{l|l|cc|cc|c|c|c|c}
\hline
&\multirow{2}{*}{Methods} &\multicolumn{2}{c|}{\textbf{SentMT}} &\multicolumn{2}{c|}{\textbf{DocMT}} &\multicolumn{1}{c|}{\textbf{ChatMT}} &\multicolumn{1}{c|}{\textbf{PerMT}} &\multicolumn{1}{c|}{\textbf{AphMT}} &\multirow{2}{*}{Avg. score} \\ 
\cline{3-6} \cline{7-9}
&\multicolumn{1}{l|}{} & \multicolumn{1}{c}{test19} & \multicolumn{1}{c|}{test20} & \multicolumn{1}{c}{TED}& \multicolumn{1}{c|}{subtitle}  &  \multicolumn{1}{c|}{chat}      & \multicolumn{1}{c|}{personalized} & \multicolumn{1}{c|}{aphorism} \\ \hline
\multirow{3}{*}{\rm{BLEU}$\uparrow$}
& {\fontfamily{lmtt}\selectfont UMLNMT} &\textbf{30.64} &\textbf{30.75}	&\textbf{25.60}	&\textbf{30.52} &\textbf{36.93}  &\textbf{37.73}     &\textbf{43.46} &\textbf{33.66} \\
& \qquad \; \; -- Prompt  &26.07 &27.49 &24.86 &29.62&36.42     &36.14  &42.66 &31.89 \\
&Training {\fontfamily{lmtt}\selectfont UMLNMT} From Scratch &30.45	&30.46	&25.03	&29.42 &35.22  &36.11     &42.37  &32.72 \\\hline
\multirow{3}{*}{\rm{BLEURT}$\uparrow$}
&{\fontfamily{lmtt}\selectfont UMLNMT} &65.36 &64.65	&66.99	 &\textbf{65.35}&68.62  &\textbf{64.19}     &\textbf{74.60}  &\textbf{67.10}\\
&\qquad \; \; -- Prompt  &64.07 &64.03 &66.58 &64.67 &\textbf{68.79}     &63.83 &74.32  &66.61\\
&Training {\fontfamily{lmtt}\selectfont UMLNMT} From Scratch &\textbf{65.63}	&\textbf{65.06}	&\textbf{67.16}	&64.66 &68.56  &64.03      &74.31  &67.05\\
\hline
\end{tabular}}}
\caption{Ablation study. BLEU and BLEURT scores (\%) in Zh$\rightarrow$En direction on different test sets. }
\label{tbl:main_res5}\vspace{-12pt}
\end{table*}


\subsubsection{Comparison to State-of-the-art Models}
In~\autoref{mainres2}, we compare our {\fontfamily{lmtt}\selectfont UMLNMT} with state-of-the-art approaches including doc2sent++~\cite{sun-etal-2022-rethinking}, CA-MCT~\cite{liang-etal-2022-msctd}, UD-NMT~\cite{lin-etal-2021-towards}, PLUVR~\cite{fang-feng-2022-neural}. Compared with previous state-of-the-art dataset-specific models, our single model jointly trained on multiple datasets achieves competitive or better performance. In particular, on the data-limited scenarios, \emph{e.g.}, TED, chat, and personalized translation in Zh$\rightarrow$En direction, and News and test2017 of MMT in En$\rightarrow$De direction, {\fontfamily{lmtt}\selectfont UMLNMT} significantly surpasses the state-of-the-art performance by 3.60$\sim$8.12 BLEU scores. 

Though the performance of {\fontfamily{lmtt}\selectfont UMLNMT} is less satisfactory for the Europarl, our {\fontfamily{lmtt}\selectfont UMLNMT} is a single model for diverse datasets while the previous methods train dedicated models on each dataset, which means more resources are required when training and deploying models in real life. That is, if we need to support translation under $N$ scenarios, a single {\fontfamily{lmtt}\selectfont UMLNMT} is enough while it requires $N$ models with the existing method.

\begin{table}[]
\small
\centering
\begin{tabular}{@{}l|cc|cc@{}}
\hline
\multirow{2}{*}{\textbf{Models}} & \multicolumn{2}{c|}{\textbf{test19}} & \multicolumn{2}{c}{\textbf{test20}} \\ \cline{2-5} 
 & Dist-1 & Dist-2 & Dist-1 & Dist-2 \\ 
\hline
SDTM & 10.75 &51.63 & 11.06 & 51.14 \\\cdashline{1-5}[4pt/2pt]
{\fontfamily{lmtt}\selectfont UMLNMT} (SentMT) & 11.15 &53.42 & 10.94 &51.15  \\
{\fontfamily{lmtt}\selectfont UMLNMT} (DocMT)  & 10.08 &49.40 & 10.26 & 48.73 \\
{\fontfamily{lmtt}\selectfont UMLNMT} (ChatMT)  & 10.80 &52.72 & 11.00 &51.92\\
{\fontfamily{lmtt}\selectfont UMLNMT} (PerMT)  & 10.20 &52.39  & 10.14 &50.11 \\
{\fontfamily{lmtt}\selectfont UMLNMT} (AphMT)  &\textbf{12.13} &\textbf{55.90} & \textbf{12.23} &\textbf{54.91} \\
\hline
\end{tabular}
\caption{Translation diversity with different prefixed prompts (\emph{e.g.}, ``SentMT'' prompt) in Zh$\rightarrow$En direction.}
\label{diversit}\vspace{-12pt}
\end{table}


\section{Analysis}
\subsection{Ablation Study}
\label{ssec:abs}
We conduct ablation studies to investigate how well the prompt of {\fontfamily{lmtt}\selectfont UMLNMT} works. The results are shown in~\autoref{tbl:main_res5}. 

(1) When removing the prefixed prompt, \emph{i.e.}, jointly training multiple datasets without prompt, the model performance greatly degrades on all scenarios in Zh$\rightarrow$En direction. This shows the necessity of the prefixed prompt that is able to guide which translation type the model should focus on as an indicator.

(2) When training {\fontfamily{lmtt}\selectfont UMLNMT} from scratch, in general, the model performance decreases slightly compared to {\fontfamily{lmtt}\selectfont UMLNMT} trained from a pretrained NMT model. This shows that continuing training from a pretrained model may help the model effectively focus on learning to prompt. 

\begin{table}[]
\small
\centering
\scalebox{0.90}{
\setlength{\tabcolsep}{0.5mm}{
\begin{tabular}{@{}c|cccccc@{}}
\hline
\multirow{1}{*}{{Proportion}} &SDTM&{{SentMT}} & {{DocMT}} &ChatMT&PerMT&AphMT \\ \cline{2-6} 
 \hline
Top-1 &43&65  & 78 & 66 & 96 & 73 \\
Top-2 &95&118  & 140 & 116 & 144 & 115 \\
Top-3 &137&159  & 180 & 163 & 177 & 153 \\
\hline
\end{tabular}}}
\caption{The manual ranked results of the translations by using different prompts.}
\label{proporation}\vspace{-12pt}
\end{table}

\subsection{Whether it Generates Diverse and High-quality Translations?}
\label{DHQ}
To further find out whether {\fontfamily{lmtt}\selectfont UMLNMT} can generate diverse and high-quality translations, we randomly sample 200 examples from test19 and test20 sets of WMT2020 in Zh$\rightarrow$En direction. For each instance, we construct it with five types of prefixed prompts as in~\autoref{eq:prefix_x} for generating five translations. 

For \textbf{diversity}, we use the Dist-1 and Dist-2~\cite{li2016deep} to evaluate the degree of diversity for each translation. The results in~\autoref{diversit} suggest that our model with different prefixed prompts can generate diverse translations compared with the dataset-specific model (SDTM), especially with the ``AphMT'' prompt where a more novel lexicon may be generated. This proves that our prefixed prompts play a key role in guiding diverse translations. 

For \textbf{quality}, following~\citet{lin-etal-2021-towards}, we ask the linguist experts to sort these translations generated by using different prefixed prompts, and the ranked top-$k$ results are shown in~\autoref{proporation}. Besides, we ask them to annotate the main translation errors including mis-translation errors (MTE), under-translation errors (UTE), over-translation errors (OTE) and grammatical errors (GE). The statistical results of main translation errors are shown in~\autoref{translation_error}. Both of the results demonstrate that our model indeed generates better translations than the dataset-specific model (SDTM). 

We also present case studies in~\autoref{case-study} of the appendix to further show the above findings.

\subsection{Whether the Non-SentMT Prompts can Maintain the Translation Styles?}
\label{style}
To figure out whether the Non-sentMT prompts can keep their translation styles, \emph{e.g.}, \textit{does the ``PerMT'' prompt indeed help the model generate more personalized translations compared to using the ``SentMT'' prompt?} we randomly sample 200 examples from each test set including TED, chat, personalized, and aphorism in Zh$\rightarrow$En direction. Following \citet{lin-etal-2021-towards}, we employ linguist experts to sort these outputs according to the relevance between the generated translations and the document context/dialogue history/historical input/genre style. The proportion results in \autoref{fig.2} show that the Non-SentMT prompts (\emph{i.e.}, DocMT, ChatMT, PerMT, and AphMT) can keep their translation styles and prompt to generate translations more in line with their styles than the ``SentMT'' prompt in most cases, proving that our method makes better use of context/dialogue history/user traits/genre style. 

Besides, we train a text-CNN~\cite{kim-2014-convolutional} classifier to discriminate which translation (generated by using Non-SentMT and ``SentMT'' prompts) is more relevant to its style. For instance, we use the sentence training data and the personalized training data to train this binary classifier. Then, we use the classifier to judge whether the translations generated by using ``SentMT'' and ``PerMT'' prompts can be identified. It is similar to other scenes (DocMT, ChatMT, AphMT). Finally, the accuracy is 88.60\%, 93.39\%, 84.70\%, and 95.10\% for DocMT, ChatMT, PerMT, and AphMT, respectively. This also shows that our model indeed exerts the advantage of effectively incorporating the additional content by prompting, and keeps rich translation styles.

\section{Related Work}
\textbf{Prompting-based Approaches for NLP.} By adding ``hints'', the prompt learning is to make better use of pre-trained language models (PLMs)~\cite{liu-etal-2021-lexicon}. Inspired by this, many prompting methods are proposed to reformulate downstream tasks into pre-training ones to exert the advantage of PLMs~\cite{sun-etal-2022-nsp,lu2022makes,NEURIPS2020_1457c0d6,lester-etal-2021-power}. 

Another line of work aims to use prompting to unify various tasks, including this study. However, existing studies mainly focus on various discriminative tasks~\cite{lester-etal-2021-power,khashabi-etal-2020-unifiedqa,https://doi.org/10.48550/arxiv.2211.14838} rather than the generation task, \emph{e.g.}, NMT. And the most significant difference is that we focus on training a versatile model that handles multiple translation types and datasets simultaneously, which is more promising.

\paragraph{NMT.} The NMT generally includes multiple types of tasks: sentence~\cite{meng-etal-2015-encoding,vaswani2017attention,zhangetal2019bridging,meng2019dtmt,DBLP:conf/acl/ZhouMZZWS22}, document~\cite{maruf-etal-2018-contextual,zhang-etal-2018-improving,ma-etal-2020-simple}, multimodal~\cite{elliott-etal-2016-multi30k}, chat~\cite{farajian-etal-2020-findings,10003654,liang-etal-2021-modeling,liang-etal-2022-bjtu}, personalized~\cite{rabinovich-etal-2017-personalized,michel-neubig-2018-extreme,lin-etal-2021-towards}, and domain-specific~\cite{bawden-etal-2020-findings}. The existing work of each line mainly focuses on designing dataset-specific training strategies, loss objectives, model architectures, and so on. In this work, we instead focus on how to unify these different transition tasks and first propose a unified model for all of them, which is more attractive. 
\begin{table}[]
\small
\centering
\scalebox{0.85}{
\setlength{\tabcolsep}{0.5mm}{
\begin{tabular}{@{}l|cccccc@{}}
\hline
\multirow{1}{*}{\textbf{Models}} &{{BLEU$\uparrow$}} &{{BLEURT$\uparrow$}} &MTE$\downarrow$ &UTE$\downarrow$ &OTE$\downarrow$ & GE$\downarrow$\\ \cline{2-7} 
 \hline
SDTM & 27.49 & 63.40 &\textbf{7.21}  & 0.18 & 7.68 & 11.50 \\\cdashline{1-7}[4pt/2pt]
{\fontfamily{lmtt}\selectfont UMLNMT} (SentMT) & 26.34 & 62.82 &8.05  & 0.17 & 5.41 & 14.00 \\
{\fontfamily{lmtt}\selectfont UMLNMT} (DocMT) & 28.77 & 63.86 &9.10  & 0.17 & 3.04 & 14.50 \\
{\fontfamily{lmtt}\selectfont UMLNMT} (ChatMT) & 27.14 & 62.86 &8.42  & \textbf{0.05} & 4.51 & 14.00 \\
{\fontfamily{lmtt}\selectfont UMLNMT} (PerMT)& \textbf{30.75} & \textbf{64.65} &{7.64}  & {0.11} & \textbf{1.78} & {14.50} \\
{\fontfamily{lmtt}\selectfont UMLNMT} (AphMT) & 28.24 & {64.42} & {8.49} & {0.09} & {2.12} & {14.50} \\ 
\hline
\end{tabular}}}
\caption{Automatic results in terms of BLEU (\%) and BLEURT (\%) and human evaluation results about mis-translation error (MTE; \%), under-translation error (UTE; \%), over-translation error (OTE; \%) and grammatical error (GE; \%).}
\label{translation_error}\vspace{-12pt}
\end{table}
\textbf{\begin{figure}[t]
    \centering
    \includegraphics[width=0.49\textwidth]{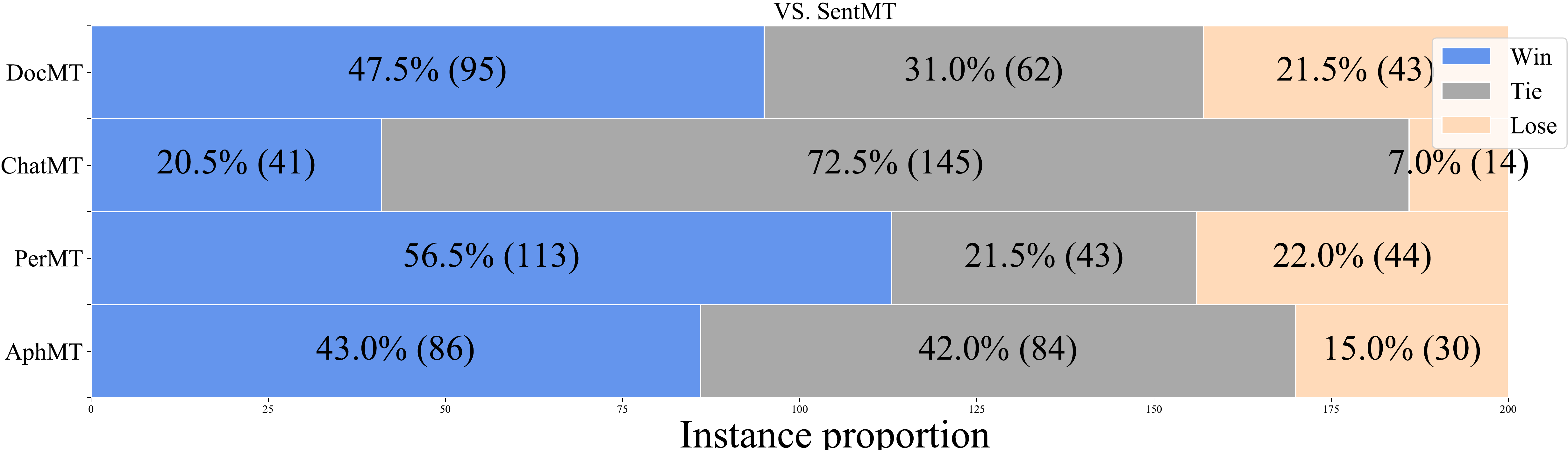}
    \caption{The proportion results among translations generated by using ``SentMT'' and non-SentMT prompts (\emph{i.e.}, SentMT vs. non-SentMT; ranked by the linguist experts). }
    \label{fig.2}\vspace{-12pt}
\end{figure}}
\section{Conclusion}
In this paper, we propose to unify multiple NMT tasks and present a versatile model that can translate well in many translation scenes. Extensive experiments in Zh$\rightarrow$En and En$\rightarrow$De directions, covering 7 types of translation tasks, show that our model achieves competitive or even better performance than state-of-the-art models in terms of automatic metrics with significantly reduced model deployment costs. Particularly, with different prefixed prompts for the same input, our model can generate diverse and high-quality translations, suggesting its superiority and generalizability.

\section*{Acknowledgements}
The research work described in this paper has been supported by the National Key R\&D Program of China (2020AAA0108001) and the National Nature Science Foundation of China (No. 61976015, 61976016, 61876198 and  61370130). 

\bibliography{anthology,custom}
\bibliographystyle{acl_natbib}
\appendix
\section*{Appendix}
\begin{table}[]
    \centering
    \resizebox{\linewidth}{!}{
    \begin{tabular}{l|c|c}
        \hline
        \multirow{1}{*}{\textbf{Hyperparameters}} & \multicolumn{1}{c|}{\textbf{Base}}  & \multicolumn{1}{c}{\textbf{Big}} \\
        \hline
        batch size & 4096 & 4096 \\
        number of GPUs & 8 & 8 \\
        hidden size & 512 & 1024 \\
        filter size & 2048 & 4096  \\
        encoder layers & 6 & 6 \\
        decoder layers & 6 & 6  \\
        attention heads & 8 & 16 \\
        residual dropout & 0.1 & 0.3 \\
        attention dropout & 0.1 & 0.1  \\
        activation dropout & 0.1 & 0.1  \\
        label smoothing & 0.1 & 0.1  \\
        learning rate & 1 & 1  \\
        warmup steps & 4000 & 8000  \\
        first-stage steps &200,000 &200,000\\
        second-stage steps &100,000 &100,000\\
        optimizer &Adam &Adam\\
        adam beta1 &0.9 &0.9\\
        adam beta2 &0.98 &0.98\\
        layer normalization & postnorm & postnorm \\
        position encoding &relative &relative\\
        share embeddings &True &True\\
        share softmax weights &False & False\\
        \hline
    \end{tabular}
    }
    \caption{Training hyperparameters and model configurations of our experiments.}
    \label{tab:train-detail}
\end{table}


\begin{table*}[]
\small
\centering
\begin{tabular}{l|p{13.8cm}l}
\hline
 Source &以上 就是 出门 的 理由 。(\textcolor{red}{The pinyin style of Chinese:} yǐshàng jiùshì chūmén de lǐyóu) \\
 \hline
  Reference &All these were the reasons to go out. \\
 \hline
Format1 &<translate the sentence> <bos> 以上 就是 出门 的 理由 。<eos> <given the context> NULL \\ \hline
Format2 &<given the context> NULL <translate the sentence> <bos> 以上 就是 出门 的 理由 。<eos>\\ \hline
Format3 &<SentMT> <bos> 以上 就是 出门 的 理由 。<eos> <context> NULL \\ \hline
\end{tabular}
\caption{The example with different formats for prompting.}
\label{prompt-example}
\end{table*}

\begin{table}[]
\small
\centering
\setlength{\tabcolsep}{0.5mm}{
\begin{tabular}{@{}c|cccc@{}}
\hline
\textbf{formats} & \textbf{SentMT (test19)} & \textbf{DocMT (TED)} & \textbf{ChatMT}  & \textbf{PerMT} \\ \hline
Format1 & 30.45 &25.34 &36.91 &37.67 \\
Format2 & 30.36 &24.89 &36.56 &37.76 \\
Format3 & 30.64 &25.60 &36.93 &37.73 \\
\hline
\end{tabular}}
\caption{The effect of using different prompt formats in Zh$\rightarrow$En direction.}
\label{prompt-matters}
\end{table}




\begin{table}[]
\small
\centering
\begin{tabular}{@{}l|cc@{}}
\hline
\multirow{2}{*}{\textbf{Models}} & \multicolumn{2}{c}{\textbf{SentMT (test20)}} \\ \cline{2-3} 
 & BLEU & BLEURT \\ 
\hline
\emph{w/o} context & 30.75 & 64.65 \\
\emph{w/ } context & 30.95 &64.99  \\
\hline
\end{tabular}
\caption{The effect of whether to use the context in Zh$\rightarrow$En direction.}
\label{context-matters}
\end{table}

\section{Implementation Details}
\label{ImD}
We list detailed training hyperparameters and model configurations for training Transformer base and big models in our experiments. We train all models using THUMT~\cite{tan-etal-2020-thumt} framework. Note that when combining all training datasets, we first filter out such instance if it appears in all test sets for preventing information leakage.

\section{Effect of Different Designs for Prompting}
In this section, we investigate the effect of different designs for prompting. The formats are shown in~\autoref{prompt-example}. The results presented in~\autoref{prompt-matters} show that these different prompt formats perform similarly in terms of BLEU scores. This suggests that adding ``hints'' for translation types (even a simple prompt) is important.

\section{Context Matters in WMT2020 News Translation}
As we all know, there is a context in the WMT20 news test set. Therefore, we investigate the effect of incorporating it. The results presented in~\autoref{context-matters} show that the context can improve the performance in terms of BLEU and BLEURT. This suggests that incorporating the context actually matters for generating better translations.

\begin{table*}[t!]
\centering
\newcommand{\tabincell}[2]{\begin{tabular}{@{}#1@{}}#2\end{tabular}}
\scalebox{0.60}{
\setlength{\tabcolsep}{0.5mm}{
\begin{tabular}{l|l|r|rrrr|r|r|r|r|r}
\hline
& &\multicolumn{1}{c|}{\textbf{SentMT}} &\multicolumn{4}{c|}{\textbf{DocMT}} &\multicolumn{1}{c|}{\textbf{ChatMT}} &\multicolumn{1}{c|}{\textbf{PerMT}} &\multicolumn{1}{c|}{\textbf{MMT}} &\multicolumn{1}{c|}{\textbf{AphMT}} &\multicolumn{1}{c}{\textbf{DsMT}} \\ 
\cline{3-6} \cline{7-12}
&\multicolumn{1}{l|}{} & \multicolumn{1}{c|}{WMT2020} & \multicolumn{1}{c}{TED} & \multicolumn{1}{c}{News} & \multicolumn{1}{c}{Europarl}& \multicolumn{1}{c|}{Subtitle} &  \multicolumn{1}{c|}{MSCTD}      & \multicolumn{1}{c|}{UTD-Corpus} & \multicolumn{1}{c|}{Multi30k}  & \multicolumn{1}{c|}{Aphorism} & \multicolumn{1}{c}{Biomedical}  \\ \hline
\multirow{3}{*}{\tabincell{c}{Zh$\rightarrow$En}}
&Training  & 22,244,006 &209,787 &  &      &2,000,000 &123,299  &14,006 &  &181,451 &    \\
&Development &2,000 &887 &  &      &2,500 &2,389  &1,557 &   &2,500 & \\
&Test & 2,000 &5,473 &  &     &2,500 &2,385  &1,536 &   &2,500 &   \\\hline
\multirow{3}{*}{\tabincell{c}{En$\rightarrow$De}}
&Train  & 45,541,367 &206,112 &236,287  &1,666,904    &  &20,240  & &29,000   & &3,035,118  \\
&Development & 1,997 &8,967 &2,169  &3,587   &   &2,674  & &1,014 &   &435 \\
&Test & 1,418 &2,271 &2,999  &5,134    &  &2,682   & &test2016: 1,000; test2017: 1,000; MSCOCO: 461 &  &505   \\
\hline
\end{tabular}}}
\caption{Dataset Statistics. For SentMT and DsMT, the development is the test19 set as we reported in the experiments. }
\label{tbl:data-statistics}\vspace{-5pt}
\end{table*}

\begin{table*}[t!]
\centering
\newcommand{\tabincell}[2]{\begin{tabular}{@{}#1@{}}#2\end{tabular}}
\begin{subtable}{.98\textwidth}
\scalebox{0.82}{
\setlength{\tabcolsep}{0.4mm}{
\begin{tabular}{l|l|cc|cc|c|c|c|cc}
\hline
&\multirow{2}{*}{Methods} &\multicolumn{2}{c|}{\textbf{SentMT}} &\multicolumn{2}{c|}{\textbf{DocMT}} &\multicolumn{1}{c|}{\textbf{ChatMT}} &\multicolumn{1}{c|}{\textbf{PerMT}} &\multicolumn{1}{c|}{\textbf{AphMT}} &\multicolumn{2}{c}{Overall} \\ 
\cline{3-6} \cline{7-11}
&\multicolumn{1}{l|}{} & \multicolumn{1}{c}{test19} & \multicolumn{1}{c|}{test20} & \multicolumn{1}{c}{TED} & \multicolumn{1}{c|}{subtitle} &  \multicolumn{1}{c|}{chat}      & \multicolumn{1}{c|}{personalized}  & \multicolumn{1}{c|}{aphorism} & \multicolumn{1}{c}{Avg.} & \multicolumn{1}{c}{\# models}  \\ \hline
\multirow{6}{*}{\tabincell{c}{ChrF2$\uparrow$}}
&Doc2Sent++~\cite{sun-etal-2022-rethinking}  & - &- &-  &-     &-  &- &- &- & 6  \\
&CA-MCT~\cite{liang-etal-2022-msctd} & - &- &-  &-  &-      &- &- &- & 6  \\
&UD-NMT~\cite{lin-etal-2021-towards}  & - &- & - &-  &-      &-  &- &- & 6  \\\cdashline{2-11}[4pt/2pt]
&SDTM &53.70 &55.08  &48.04 &30.90 &52.30 &47.12 &63.15&50.04&6\\
&{\fontfamily{lmtt}\selectfont UMLNMT} (base) &\;{59.92}$^{\dagger}$ &\;\textbf{59.72}$^{\dagger}$	&48.61	&\;{46.16}$^{\dagger}$ &52.32  &\;{63.73}$^{\dagger}$      &63.44 &56.27 &1\\
&\;\;\;\; -- Prompt  &55.50 &55.08 &48.25 &46.37 &52.06     &62.58  &62.87 &54.67 &1\\

&{\fontfamily{lmtt}\selectfont UMLNMT} (big)  &\;\textbf{60.28}$^{\dagger}$  &\;{59.41}$^{\dagger}$ &\;\textbf{48.94}$^{\dagger}$ &\;\textbf{47.38}$^{\dagger}$ &\textbf{52.69} &\;\textbf{65.46}$^{\dagger}$    &\;\textbf{63.74}$^{\dagger}$&\textbf{56.84}&1\\ 
\hline
\multirow{7}{*}{\tabincell{c}{TER$\downarrow$}}
&Doc2Sent++~\cite{sun-etal-2022-rethinking}  & - &- &-  &-     &-  &- &- &- & 6  \\
&CA-MCT~\cite{liang-etal-2022-msctd} & - &- &- &-  &51.06  &-      &- &- & 6  \\
&UD-NMT~\cite{lin-etal-2021-towards}  & - &- & - &-  &-      &-  &- &- & 6  \\\cdashline{2-11}[4pt/2pt]
&SDTM &57.62 &56.91  &57.27 &\textbf{50.22} &46.06 &46.92  &37.50&50.36&6\\
&{\fontfamily{lmtt}\selectfont UMLNMT} (base) &\;{52.13}$^{\dagger}$ &\;{54.64}$^{\dagger}$	&57.51	&50.43 &46.06  &46.42      &\;{36.27}$^{\dagger}$ &49.07 &1\\
&\;\;\;\; -- Prompt  &55.47 &56.91 &57.44  &51.16 &46.37     &47.00 &37.06  &50.20&1\\
&{\fontfamily{lmtt}\selectfont UMLNMT} (big)  &\;\textbf{51.73}$^{\dagger}$  &\;\textbf{53.56}$^{\dagger}$ &\;\textbf{57.36}$^{\dagger}$ &{50.29} &\textbf{45.76} &\;\textbf{43.66}$^{\dagger}$    &\;\textbf{36.08}$^{\dagger}$&\textbf{48.35}&1\\ \hline
\multirow{7}{*}{\tabincell{c}{COMET$\uparrow$}}
&Doc2Sent++~\cite{sun-etal-2022-rethinking}  & - &- &-  &-     &-  &- &- &- & 6  \\
&CA-MCT~\cite{liang-etal-2022-msctd} & - &- &-  &-  &-      &- &- &- & 6  \\
&UD-NMT~\cite{lin-etal-2021-towards}  & - &- & - &-  &-      &-  &- &- & 6  \\\cdashline{2-11}[4pt/2pt]
&SDTM &24.76 &26.22  &39.09  &32.14 &47.01 &30.83 &68.37&38.35&6\\
&{\fontfamily{lmtt}\selectfont UMLNMT} (base) &\;{34.21}$^{\dagger}$ &\;{33.62}$^{\dagger}$	&\;{41.62}$^{\dagger}$	&\;{33.51}$^{\dagger}$  &47.01  &\;{31.74}$^{\dagger}$     &\;{70.18}$^{\dagger}$  &41.70&1 \\
&\;\;\;\; -- Prompt   &28.77 &30.03 &40.74 &31.65 &45.97    &29.54  &69.38 &39.44&1\\
&{\fontfamily{lmtt}\selectfont UMLNMT} (big)  &\;\textbf{38.46}$^{\dagger}$  &\;\textbf{38.13}$^{\dagger}$ &\;\textbf{43.33}$^{\dagger}$  &\;\textbf{33.27}$^{\dagger}$ &\;\textbf{48.10}$^{\dagger}$ &\;\textbf{37.56}$^{\dagger}$    &\;\textbf{70.71}$^{\dagger}$&\textbf{44.22}&1\\ \hline
\end{tabular}}}
\caption{Zh$\rightarrow$En. ``$^{\dagger}$'' indicates that statistically significantly better than the ``SDTM'' with t-test {\em p} \textless \ 0.01. }
\end{subtable}
\centering

\begin{subtable}{.98\textwidth}
\scalebox{0.66}{
\setlength{\tabcolsep}{0.4mm}{
\begin{tabular}{l|l|cc|ccc|ccc|c|cc|cc}
\hline
&\multirow{2}{*}{Methods} &\multicolumn{2}{c|}{\textbf{SentMT}} &\multicolumn{3}{c|}{\textbf{DocMT}} &\multicolumn{3}{c|}{\textbf{MMT}} &\multicolumn{1}{c|}{\textbf{ChatMT}} &\multicolumn{2}{c|}{\textbf{DsMT}} &\multicolumn{2}{c}{Overall} \\ 
\cline{3-15} 
&\multicolumn{1}{l|}{} & \multicolumn{1}{l}{test19} & \multicolumn{1}{l|}{test20} & \multicolumn{1}{c}{Europarl} & \multicolumn{1}{c}{News} & \multicolumn{1}{c|}{TED} &  \multicolumn{1}{c}{test2016}      & \multicolumn{1}{c}{test2017} & \multicolumn{1}{c|}{MSCOCO} & \multicolumn{1}{c|}{chat}& \multicolumn{1}{c}{test19} & \multicolumn{1}{c|}{test20} & \multicolumn{1}{c}{Avg.} & \multicolumn{1}{c}{\# models}\\ \hline
\multirow{7}{*}{\tabincell{c}{ChrF2$\uparrow$}}
&Doc2Sent++~\cite{sun-etal-2022-rethinking}   &- & -& -  & - & - & - & - & - & - & - & -& - &7\\
&PLUVR~\cite{fang-feng-2022-neural}  &- & -& -  & - & - & - & - & - & - & - & -& - &7\\
&CA-MCT~\cite{liang-etal-2022-msctd}   &- & -& -  & - & - & - & - & - & - & - & - & - &7\\\cdashline{2-15}[4pt/2pt]
&SDTM  &65.28 & 59.34  & 60.29  & 61.37& 57.19  & 67.05  & 65.13 & \textbf{60.94} & 69.29 & 59.71 & 56.94 &62.05&7\\
&{\fontfamily{lmtt}\selectfont UMLNMT} (base) & 65.84 & 60.88 & 60.43 & 62.83  & 58.38  & 67.10 & 65.67 & 59.78 & 67.77 &\;\textbf{60.61}$^{\dagger}$  &\;\textbf{58.75}$^{\dagger}$&{62.54}&1\\
&\;\;\;\; -- Prompt  & 64.52 & 57.40  & 60.30 & 62.37  & 58.29  & 67.44 & 65.08 & 58.39 & 66.23  & 59.36  & 56.53 &61.45&1\\
& {\fontfamily{lmtt}\selectfont UMLNMT} (big)  &\;\textbf{66.71}$^{\dagger}$  &\;\textbf{61.55}$^{\dagger}$ &\textbf{60.65}  &\;\textbf{63.10}$^{\dagger}$ &\;\textbf{59.02}$^{\dagger}$  &\;\textbf{67.94}$^{\dagger}$  &\;\textbf{66.64}$^{\dagger}$&60.09&\textbf{69.35}&59.47&57.18&\textbf{62.91}&1\\ 
\hline
\multirow{7}{*}{\tabincell{c}{TER$\downarrow$}}
&Doc2Sent++~\cite{sun-etal-2022-rethinking}   &- & -& -  & - & - & - & - & - & - & - & -& - &7\\
&PLUVR~\cite{fang-feng-2022-neural}  &- & -& -  & - & - & - & - & - & - & - & -& - &7\\
&CA-MCT~\cite{liang-etal-2022-msctd}   &- & -& -  & - & - & - & - & - &29.39 & - & - & - &7\\\cdashline{2-15}[4pt/2pt]
&SDTM  &42.09 & 49.52& 53.48  & 56.16 & 53.99 & 38.56 & 43.93 & \textbf{46.86} & 28.47 & 55.36 & 55.89&47.66&7\\
&{\fontfamily{lmtt}\selectfont UMLNMT} (base) & 42.57 	&  48.74 &  53.40 	&  46.01  &  53.54     & 38.23  &  42.94  & 49.46 &  29.71 & \;\textbf{53.94}$^{\dagger}$  &  \;\textbf{53.36}$^{\dagger}$ &\textbf{46.54}&1\\
&\;\;\;\; -- Prompt  & 42.40  & 51.25 & 53.60 & 47.38  & 54.38 & 40.59 & 43.78 & 54.40& 32.98  & 55.89 & 57.43 &48.55&1\\
&{\fontfamily{lmtt}\selectfont UMLNMT} (big)  &\textbf{42.01}  &\;\textbf{47.55}$^{\dagger}$ &\textbf{53.08} &\;\textbf{44.94}$^{\dagger}$ &\;\textbf{52.85}$^{\dagger}$  &\;\textbf{36.87}$^{\dagger}$  &\;\textbf{42.58}$^{\dagger}$&54.98&\textbf{28.45}&54.42&55.28&46.63&1\\ 
\hline
\multirow{7}{*}{\tabincell{c}{COMET$\uparrow$}}
&Doc2Sent++~\cite{sun-etal-2022-rethinking}   &- & -& -  & - & - & - & - & - & - & - & -& - &7\\
&PLUVR~\cite{fang-feng-2022-neural}  &- & -& -  & - & - & - & - & - & - & - & -& - &7\\
&CA-MCT~\cite{liang-etal-2022-msctd}   &- & -& -  & - & - & - & - & - & - & - & - & - &7\\\cdashline{2-15}[4pt/2pt]
&SDTM  & 40.68 &  33.62 & 59.54  & 51.82 & 46.19 & 59.67 & 52.24 & 39.93& 62.80& 35.19& 31.12&46.62&7\\
&{\fontfamily{lmtt}\selectfont UMLNMT} (base) & \;47.12$^{\dagger}$	& \;38.83$^{\dagger}$	& 59.72	& \;53.47$^{\dagger}$  & \;47.84$^{\dagger}$    & \;62.08$^{\dagger}$ & \;57.36$^{\dagger}$ & 37.55& 61.07& \;\textbf{67.19}$^{\dagger}$ & \;\textbf{62.49}$^{\dagger}$&54.07&1\\
&\;\;\;\; -- Prompt  & 40.85 & 30.50  & 59.02 & 53.21    & 46.27 & 60.49& 55.38& 32.32& 56.58 & 42.63 & 39.64&46.99&1\\
&{\fontfamily{lmtt}\selectfont UMLNMT} (big)  &\;\textbf{49.62}$^{\dagger}$  &\;\textbf{40.72}$^{\dagger}$ &\textbf{60.21}  &\;\textbf{55.38}$^{\dagger}$ &\;\textbf{49.11}$^{\dagger}$  &\;\textbf{63.47}$^{\dagger}$  &\;\textbf{60.90}$^{\dagger}$&\textbf{40.09}&\textbf{63.55}&\;65.30$^{\dagger}$&\;61.44$^{\dagger}$&\textbf{55.43}&1\\ 
\hline
\end{tabular}}}
\caption{En$\rightarrow$De. ``$^{\dagger}$'' indicates that statistically significantly better than the ``SDTM'' with t-test {\em p} \textless \ 0.01}
\end{subtable}
\caption{ChrF2, TER, and COMET scores (\%) on test sets. ``-'' indicates no such result in the original paper.  }
\label{tbl:main-res-others}\vspace{-10pt}
\end{table*}

\section{Case Study}
\label{CS}

In this section, we random sample several instances from each test set and present them in~\autoref{case-study} to give some observations among the single-dataset training model (SDTM) and ours using different prefixed prompts.

(1) Compared to the SDTM, we observe that our model translations are always better. This indicates that our model indeed benefits from different tasks and learns more information from different datasets. 

(2) For all case~\autoref{case-study}, it is easy to find that our translations generated by different prefixed prompts are diverse and high-quality, as shown in~\autoref{DHQ}. Even for a simple case, \emph{e.g.},~\autoref{case-study} (a), our model still produces meaningful translations with a rich lexicon, which is able to serve well as an important augmented method for NMT. This suggests that the proposed promoting approach is effective and indispensable for a successful and `intelligent' translator, which can meet the increasing demand of various users and enhance the diversity of the augmented corpus.

(3) In particular, for the case~\autoref{case-study} (c), we find that our method with two prompts (\emph{i.e.,}, DocMT and ChatMT) translate the entity accurately, ``炮铜 (pàotóng)'' while others fail. The reason may be that both prompts can fully understand the dialogue context while other prompts cannot because they have no such chance to access the context during training. This also shows that our Non-SentMT prompts can maintain their translation styles as we demonstrated in~\autoref{style}. 

In summary, all cases show that our {\fontfamily{lmtt}\selectfont UMLNMT} model enhanced by the proposed perfixed prompts and training manner yields diverse and satisfactory translations, showing its superiority and generalizability.

\begin{table*}[]
\small
\centering
\begin{subtable}{.98\textwidth}
\centering
\scalebox{0.86}{
\begin{tabular}{l|p{13.8cm}l}
\hline
 Source &以上 就是 出门 的 理由 。(\textcolor{red}{The pinyin style of Chinese:} yǐshàng jiùshì chūmén de lǐyóu) \\
 \hline
  Reference &all these were the reasons to go out. \\
 \hline
SDTM & that's why we're going out.\\ \hline
{\fontfamily{lmtt}\selectfont UMLNMT} (SentMT) & these are the reasons for going out.\\ \hline
{\fontfamily{lmtt}\selectfont UMLNMT} (DocMT) & that's why you go out. \\ \hline
{\fontfamily{lmtt}\selectfont UMLNMT} (ChatMT) & that's why we're going out.  \\ \hline
{\fontfamily{lmtt}\selectfont UMLNMT} (PerMT) & above is the reason for going out。 \\ \hline
{\fontfamily{lmtt}\selectfont UMLNMT} (AphMT) & that's the reason to go out. \\ 
\hline
\end{tabular}}
\caption{The input sentence example comes from sentence translation (test20).}
\end{subtable}
\begin{subtable}{.98\textwidth}
\centering
\scalebox{0.86}{
\begin{tabular}{l|p{13.8cm}l}
\hline
 \multirow{2}{*}{Source} &我 外婆 坐在 房间 的 另 一 端 盯着 我看 。 (\textcolor{red}{The pinyin style of Chinese:} wǒ wàipó zuòzài fángjiān de lìng yī duān dīngzhe wǒkàn） \\
 \hline
  Reference &and my grandmother was sitting across the room staring at me. \\
 \hline
 \multirow{3}{*}{\textcolor[RGB]{0,176,80}{Document context}} & 我的 表兄妹们 总是 无处不在 [SEP] 我 记得 ， 当 我 八九 岁 时 的 一 次 我 早上 醒来 ， 跑到 客厅 所有 的 表兄妹 都 在 (\textcolor{red}{The pinyin style of Chinese:} wǒde biǎoxiōngmèimen zǒngshì wúchùbúzài [SEP] wǒ jìde, dāng wǒ bājiǔ suì shí de yī cì wǒ zǎoshang xǐnglái, pǎodào kètīng suǒyǒu de biǎoxiōngmèi dōu zài))\\
 \hline
SDTM & and my grandmother sat across the room staring at me. \\ \hline
{\fontfamily{lmtt}\selectfont UMLNMT} (SentMT) & and my grandmother sat on the other side of the room staring at me. \\ \hline
{\fontfamily{lmtt}\selectfont UMLNMT} (DocMT) & my grandmother was sitting on the other side of the room staring at me. \\ \hline
{\fontfamily{lmtt}\selectfont UMLNMT} (ChatMT) & and my grandmother was sitting at the other end of the room staring at me. \\ \hline
{\fontfamily{lmtt}\selectfont UMLNMT} (PerMT) & my grandmother sat at the other end of the room staring at me. \\ \hline
{\fontfamily{lmtt}\selectfont UMLNMT} (AphMT) & my grandmother sat at the other end of the room staring at me my cousins were always everywhere.\\ 
\hline
\end{tabular}}
\caption{The input sentence example comes from the document-level translation (TED).}
\end{subtable}
\begin{subtable}{.98\textwidth}
\centering
\scalebox{0.86}{
\begin{tabular}{l|p{13.8cm}l}
\hline
 \multirow{2}{*}{Source} &重复 ， 呼叫 ， 雷霆 ， 我 是 炮铜 2-5 ， 完毕 。(\textcolor{red}{The pinyin style of Chinese:} chóngfù, hūjiào, léitíng, wǒ shì pàotóng 2-5, wánbì) \\
 \hline
  Reference & i repeat, thunder ops, this is \textcolor[RGB]{30,144,255}{\uline{gunmetal}} 2-5. over. \\
 \hline
\multirow{2}{*}{\textcolor[RGB]{0,176,80}{Dialogue history}} & 我们 需要 尽快 离开 山谷 。 [SEP] thunder ops, this is \textcolor[RGB]{30,144,255}{\uline{gunmetal}} 2-5. over. (\textcolor{red}{The pinyin style of Chinese:} wǒmen xūyào jǐnkuài líkāi shāngǔ.)\\
 \hline
SDTM & repeat, call, thunder, this is gunpowder 2-5. over.\\ \hline
{\fontfamily{lmtt}\selectfont UMLNMT} (SentMT) & repeat, thunder, this is gunpowder 2-5. over.\\ \hline
{\fontfamily{lmtt}\selectfont UMLNMT} (DocMT) & repeat, call, thunder, this is \textcolor[RGB]{30,144,255}{\uline{gunmetal}} 2-5, over. \\ \hline
{\fontfamily{lmtt}\selectfont UMLNMT} (ChatMT) & repeat, call, thunder, this is \textcolor[RGB]{30,144,255}{\uline{gunmetal}} 2-5, over. \\ \hline
{\fontfamily{lmtt}\selectfont UMLNMT} (PerMT) & repeat, call, thunder, this is gunpowder 2-5, over. \\ \hline
{\fontfamily{lmtt}\selectfont UMLNMT} (AphMT) & repeat, call, thunder, this is gunpowder 2-5, over.\\ 
\hline
\end{tabular}}
\caption{The input sentence example comes from the chat translation.}
\end{subtable}
\begin{subtable}{.98\textwidth}
\centering
\scalebox{0.86}{
\begin{tabular}{l|p{13.8cm}l}
\hline
 \multirow{3}{*}{Source} &2020 欧美 秋季 新款 女装 v领 性感 挂脖 绑带 背心 开叉 长裙 两件套 套装 女 (\textcolor{red}{The pinyin style of Chinese:} 2020 ōuměi qiūjì xīnkuǎn nǚzhuāng vlǐng xìnggǎn guàbó bǎngdài bèixīn kāichā chángqún liǎngjiàntào tàozhuāng nǚ) \\
 \hline
  \multirow{2}{*}{Reference} &2020 european and american new fall women 's clothing v-neck sexy halter strap vest split dress two-piece suit for women \\
 \hline
\multirow{3}{*}{\textcolor[RGB]{0,176,80}{Historical inputs}} &不 发货哒 固定 指定 片 花色 随机 两 件 套套 装 女 女装 开叉 性感 挂脖 新款 欧 美 秋季 绑带 背心 长裙 (\textcolor{red}{The pinyin style of Chinese:} bù fāhuòdā gùdìng zhǐdìng piàn huāsè suíjī liǎng jiàn tàotào zhuāng nǚ nǚzhuāng kāichā xìnggǎn guàbó xīnkuǎn ōu měi qiūjì bǎngdài bèixīn chángqún)\\
 \hline
\multirow{2}{*}{SDTM} &  2020 european and american autumn new women 's v-neck sexy scarf strap vest split skirt two-piece suit women\\ \hline
{\fontfamily{lmtt}\selectfont UMLNMT} (SentMT) & 2020 european and american autumn new women 's v-neck sexy neck strap vest split long skirt two-piece suit\\ \hline
\multirow{2}{*}{{\fontfamily{lmtt}\selectfont UMLNMT} (DocMT)} & 2020 european and american autumn new women ' s v-neck sexy neck strap vest split skirt two-piece suit women \\ \hline
\multirow{2}{*}{{\fontfamily{lmtt}\selectfont UMLNMT} (ChatMT)} & 2020 european and american autumn new women 's v-neck sexy scarf strap vest split long skirt two-piece suit women \\ \hline
\multirow{2}{*}{{\fontfamily{lmtt}\selectfont UMLNMT} (PerMT)} & 2020 european and american autumn new women ' s v-neck sexy neck strapping vest split long skirt two pieces set women \\ \hline
\multirow{2}{*}{{\fontfamily{lmtt}\selectfont UMLNMT} (AphMT)} & 2020 european and american autumn new women 's v-neck sexy hanging neck strap vest open fork long skirt two-piece suit women 's undelivered \\ 
\hline
\end{tabular}}
\caption{The input sentence example comes from the personalized translation.}
\end{subtable}
\begin{subtable}{.98\textwidth}
\centering
\scalebox{0.86}{
\begin{tabular}{l|p{13.8cm}l}
\hline
\multirow{3}{*}{Source} & 除了 你 ， 没有 人 能 控制 你 的 幸福 ； 因此 ， 你 有 能力 改变 你 自己 或 你 的 生活 中 任何 你 想 改变 的 东西 。(\textcolor{red}{The pinyin style of Chinese:} chúle nǐ, méiyǒu rén néng kòngzhì nǐ de xìngfú; yīncǐ, nǐ yǒu nénglì gǎibiàn nǐ zìjǐ huò nǐ de shēnghuó zhōng rènhé nǐ xiǎng gǎibiàn de dōngxī.) \\
 \hline
  \multirow{2}{*}{Reference} &no one is in control of your happiness but you; therefore, you have the power to change anything about yourself or your life that you want to change. \\
 \hline
\multirow{2}{*}{SDTM} & no one can control your happiness except you; therefore, you have the power to change yourself or anything in your life that you want to change .\\ \hline
\multirow{2}{*}{{\fontfamily{lmtt}\selectfont UMLNMT} (SentMT)} & no one can control your happiness except you; therefore, you have the ability to change yourself or anything in your life that you want to change. \\ \hline
\multirow{2}{*}{{\fontfamily{lmtt}\selectfont UMLNMT} (DocMT)} & no one can control your happiness except you; therefore, you have the power to change yourself or anything in your life that you want to change. \\ \hline
\multirow{2}{*}{{\fontfamily{lmtt}\selectfont UMLNMT} (ChatMT)} & no one can control your happiness but you; therefore, you have the power to change yourself or anything in your life that you want to change. \\ \hline
\multirow{2}{*}{{\fontfamily{lmtt}\selectfont UMLNMT} (PerMT)} &  no one can control your happiness except you; therefore, you have the ability to change yourself or anything in your life you want to change. \\ \hline
\multirow{2}{*}{{\fontfamily{lmtt}\selectfont UMLNMT} (AphMT)} & no one can control your happiness but you; therefore, you have the ability to change yourself or anything in your life you want to change.\\ 
\hline
\end{tabular}}
\caption{The input sentence example comes from aphorism translation.}
\end{subtable}
\caption{The model outputs using different prefixed prompts for each random sample of the corresponding test set. }
\label{case-study}
\end{table*}

\end{CJK}
\end{document}